\documentclass[a4paper,twocolumn]{article}

\usepackage{graphicx}
\usepackage{amsmath}      
\usepackage{amssymb}      
\usepackage{url}          
\usepackage{hyperref}     
\usepackage{booktabs}     
\usepackage{multirow}     
\usepackage{float}        

\title{Context-Aware Palmprint Recognition via a Relative Similarity Metric}

\author{
Trinnhallen Brisley, \\
                University of Edinburgh \\
                t.brisley@sms.ed.ac.uk
\and
Joseph Magen \\ Nethermind \\           
                joseph.magen@nethermind.io
\and
Aryan Gandhi \\ University of Waterloo, Nethermind \\
                aryan.gandhi@nethermind.io
}

\makeatletter
\long\def\@copyrightspace{}
\makeatother

\begin{document}
\maketitle

\begin{abstract}
We propose a new approach to matching mechanism for palmprint recognition by introducing a Relative Similarity Metric (RSM) that enhances the robustness and discriminability of existing matching frameworks. While conventional systems rely on direct pairwise similarity measures—such as cosine or Euclidean distances—these metrics fail to capture how a pairwise similarity compares within the context of the entire dataset. Our method addresses this by evaluating the relative consistency of similarity scores across all identities, allowing for better suppression of false positives and negatives. Applied atop the state-of-the-art CCNet architecture, our method achieves a new state-of-the-art 0.000036\% Equal Error Rate (EER) on the Tongji dataset, outperforming previous methods and demonstrating the efficacy of incorporating relational structure into the palmprint matching process.
\end{abstract}

\section{Introduction}

Palmprint recognition has emerged as a reliable and discriminative biometric modality due to the richness of features it offers—ranging from principal lines and creases to fine-grained textures and wrinkles. Traditional palmprint recognition systems typically involve extracting handcrafted or learned features, followed by computing pairwise similarities using Euclidean, cosine, or Hamming distance metrics. While these distance-based methods have shown strong baseline performance, they evaluate similarity scores between image pairs in isolation, ignoring the broader relational structure within the dataset.

With the rise of deep learning, several advanced architectures have been proposed to improve representation learning and matching. Svoboda et al.~\cite{svoboda2016palmprint} introduced a discriminative index learning framework using a d-prime optimized CNN, effectively increasing the separation between genuine and impostor scores. Liu et al.~\cite{liu2023smhnet} presented SMHNet, a similarity-preserving hashing network tailored for few-shot palmprint recognition that maps features to binary space for efficient matching and storage.

Despite these advances, most recognition systems still rely on direct pairwise comparisons between two images without considering how those similarities behave in the context of other images in the dataset. To address this, we propose a Relative Similarity Metric (RSM), which evaluates the similarity between a pair of palmprints relative to the similarity patterns formed between one of those images and all others in the dataset. This approach is motivated by the hypothesis that samples from the same identity should generate consistent similarity patterns with respect to a third image—a notion inspired by recent findings in fingerprint biometrics by Tang et al.~\cite{tang2024intrafinger}, which demonstrated latent similarity structures even across different fingers of the same subject.

Our method integrates this contextual similarity notion into the matching phase of a state-of-the-art hashing-based system. While traditional models such as CCNet use cosine similarity for verification, our approach enhances CCNet by evaluating similarity in a relational context, offering a more robust and interpretable basis for decision-making.

We show that this formulation leads to improved match separation, reduction of outlier behavior in impostor and genuine distributions, and ultimately a drop in Equal Error Rate (EER). By leveraging this relative structure, our work opens a new direction in biometric comparison that moves beyond isolated similarity scoring toward context-aware relational reasoning.

\section{Palmprint Matching Algorithms}

Palmprint recognition techniques have evolved significantly from handcrafted feature extraction to deep learning and hashing-based approaches. Early palmprint systems focused on texture-based descriptors such as principal lines, Gabor filters, and local binary patterns (LBP) \cite{kong2006competitive, zhang2003online, wu2004palmprint}. These descriptors were followed by traditional classification algorithms, including Support Vector Machines (SVMs), k-Nearest Neighbors (k-NN), and Linear Discriminant Analysis (LDA), to match palmprint identities.

Classification-based approaches gained a boost with the introduction of deep learning. Convolutional Neural Networks (CNNs) enabled end-to-end learning of features directly from raw palmprint images, improving recognition performance under varying lighting, orientation, and occlusion conditions. The PalmNet architecture \cite{zhang2020palmnet} is a notable example that introduced a deep network specifically for palmprint identification and verification. This was followed by variants such as Deep Distillation Hashing (DDH) \cite{deepdistillation2021}, which employed a teacher-student learning strategy to bridge the gap between constrained and unconstrained palmprint recognition. Similarly, CCNet \cite{ccnet2023} uses a multi-branch architecture with attention-based channel competition to extract robust features, improving both intra-class compactness and inter-class separability.

Zhong et al.~\cite{zhong2019handgraph} proposed a hand-based multi-biometric system using deep hashing and graph matching, while their earlier work \cite{zhong2019survey} offers a comprehensive review of palmprint advancements over the last decade. Zhang et al.~\cite{zhang2018lbp} enhanced classification accuracy by integrating modified LBP with weighted Sparse Representation-based Classification (SRC). Fei et al.~\cite{fei2022jointly} introduced a joint discriminative feature learning strategy, while their earlier work on 3D palmprints \cite{fei20203dpalm} explored feature extraction from depth-enhanced modalities. Other relevant CNN-based systems include PalmNet-Gabor \cite{genovese2019palmnet}, Learnable Gabor Convolution Networks (LGCN) \cite{chen2019lgcn}, and deep open-set identification via centralized cosine loss \cite{zhong2020cosine}.

In terms of hashing, several frameworks convert learned features into compact binary codes for efficient large-scale matching. DHN \cite{zhu2016dhn}, DHPN \cite{deepdistillation2021}, and DDH \cite{deepdistillation2021} are representative examples. These models rely on similarity-preserving loss functions, such as contrastive loss \cite{hermans2017defense} and supervised contrastive learning \cite{khosla2020supervised}, to enforce semantic consistency in Hamming space. SMHNet \cite{liu2023smhnet} further introduces a similarity metric hashing strategy for few-shot palmprint scenarios.

Numerous metric learning approaches have been explored to enhance binary code separability. These include half-orientation feature extraction \cite{fei2016half}, adaptive feature selection \cite{li2015feature}, and co-learning strategies for deployment on IoT platforms \cite{dong2022cohash}. Wu et al.~\cite{wu2021fusion} fused palmprint and palmvein modalities via a deep hashing network, demonstrating robustness in multimodal biometrics. Additional approaches include the use of rank-based and margin-based losses, such as in \cite{zhong2020cosine}, to increase discriminative power of learned hash codes.

Texture and coding-based techniques have also been extended in recent years. For example, Binary Orientation Co-occurrence Vector (BOCV) \cite{guo2009bocv} and its enhanced versions E-BOCV \cite{yang2021extreme} achieved competitive results by encoding local orientation information. Multi-order texture coding methods \cite{yang2023multiorder} further exploit structural patterns across palmprint regions. Fusion methods at the feature level \cite{kong2006fusion} and hybrid classification-retrieval systems \cite{zhang2021comprehensive} have also proven effective.

An equally critical component in palmprint recognition systems is the matching mechanism, which determines how similarity between features is evaluated. Traditional classifiers often rely on hard decision boundaries and predicted labels for identity classification. In contrast, modern systems increasingly leverage continuous similarity scores, enabling ranking-based evaluation and threshold tuning for verification tasks.

Cosine similarity has become a widely adopted measure, particularly in deep embedding-based models. It computes the cosine of the angle between two feature vectors and is particularly effective in measuring orientation-based alignment, regardless of vector magnitude. Methods such as CCNet \cite{ccnet2023} utilize cosine similarity directly in the matching stage to compare feature vectors extracted from two palmprint images. This approach allows for soft decisions and robust handling of intra-class variation, making it suitable for open-set scenarios.

Euclidean distance remains another common choice, especially during training with loss functions such as contrastive loss and triplet loss. These losses encourage smaller Euclidean distances for matching pairs and larger distances for non-matching pairs, shaping the embedding space to reflect semantic proximity. While slightly more computationally intensive than cosine similarity, Euclidean distance offers a geometrically interpretable metric that is often aligned with the learned feature distributions.

In hashing-based systems, Hamming distance is the predominant metric used at inference time due to its computational efficiency. Once continuous feature embeddings are binarized (e.g., via a \texttt{sign} or \texttt{tanh} operation), Hamming distance simply counts the number of differing bits between two binary codes. This allows for rapid large-scale matching using bitwise XOR operations. Deep Hashing Networks (DHN) \cite{zhu2016dhn} and Deep Distillation Hashing (DDH) \cite{deepdistillation2021} rely heavily on this property to achieve real-time performance in identification tasks.

Some classification-based approaches, such as PalmNet \cite{zhang2020palmnet}, avoid pairwise matching altogether during testing by directly predicting identity labels via a softmax layer. This approach is well-suited for closed-set identification but lacks flexibility in unseen identity generalization or verification settings.

Overall, the choice of matching mechanism is intimately tied to the network architecture, loss function, and target application (e.g., identification vs. verification). While direct pairwise similarity measures like cosine and Euclidean distances offer continuous and interpretable scores, Hamming distance excels in scalability. Our proposed Relative Similarity Metric (RSM) builds on these foundations by introducing a context-aware comparison framework that evaluates similarity not in isolation but in relation to the broader dataset.

Datasets like Tongji \cite{zhang2017towards} have driven benchmarking efforts. The dataset includes 12,000 contactless palmprint images from 300 individuals (600 palms), captured across two sessions. Each palm contributes 10 images per session, yielding 20 images per subject. Genuine and impostor matching totals reach 60,000 and 35,940,000, respectively, making the dataset ideal for evaluating both intra- and inter-session variation under extreme class imbalance.

In summary, palmprint recognition continues to advance through increasingly sophisticated models, moving beyond isolated pairwise comparisons to incorporate context-aware and relational similarity metrics, as explored in this work through the Relative Similarity Metric (RSM).

\section{Relative Similarity Metric}

Traditional palmprint recognition systems typically rely on absolute similarity measures between image pairs. These approaches, while effective, fail to account for how a pairwise similarity compares to other similarities involving the same reference image. We introduce a Relative Similarity Metric (RSM) to address this limitation. RSM provides a higher-order measure of consistency in similarity behavior by evaluating not only a given similarity score but also how it contrasts with scores involving all other images in the dataset.

Let \( \mathcal{D} = \{ (x_i, y_i) \}_{i=1}^{N} \) denote the dataset, where \( x_i \) is the feature representation (e.g., a hash code or embedding) of image \( i \), and \( y_i \in \{1, \dots, C\} \) is the identity label (user) associated with that image. Let \( S(x_i, x_j) \) be a similarity function, such as cosine distance or the Hamming distance, such that a similarity score of 0 implies perfect similarity and a score of 1 implies perfect dissimilarity. Thus, the higher the value of \( S(x_i, x_j) \), the more dissimilar the images \(x_i \) and \(x_j \) are from one another.

We define the Relative Similarity Metric with respect to a reference image \( x_k \in \mathcal{D} \). For a fixed pair of images \( x_i\) and \( x_j \), we compute the similarity between \( x_i \) and \( x_k \), as well as between \( x_j \) and \( x_k \). The relative difference between these similarities is given by:

\begin{equation}
\Delta_{i,j}^{(k)} = |S(x_i, x_k) - S(x_k, x_j)|,
\label{eq:delta}
\end{equation}

This measures how comparatively similar images \( x_i \) and \( x_j \) relate to the reference image \( x_k \). For each reference \( x_k \), we group \( x_k \in \mathcal{D} \setminus \{x_i\} \) based on identity agreement with \( x_i \) using identity labels. Specifically:

\begin{align}
\mathcal{M}_i^{(j)} &= \{ x_k \in \mathcal{D} \setminus \{x_i\} \mid y_k = y_i \}, \\
\mathcal{N}_i^{(j)} &= \{ x_k \in \mathcal{D} \setminus \{x_i\} \mid y_k \neq y_i \},
\end{align}

corresponding to images that share or do not share identity with \( x_i \), respectively. These identity-based sets are used for analysis and interpretation but are not required during inference.

The relative similarity score of \( x_i \) with respect to \( x_j \) is defined as the similarity between \( x_i \) and \( x_j \) and the average similarity of other sample images to \( x_i \) and \(x_j \):

\begin{equation}
\begin{aligned}
R(x_i, x_j) = \frac{1}{M} \sum_{x_k \in \mathcal{D'} \setminus \{x_i, x_j\}} \Delta_{i,k}^{(j)} + \alpha S(x_i, x_j).
\end{aligned}
\label{eq:rsm_score}
\end{equation}

Here, $\mathcal{D'}$ denotes a random sample of $M$ images drawn from the database. That is, $M$ corresponds to the number of reference image comparisons used to compute the metric. Increasing $M$ typically improves the stability and accuracy of the RSM score, albeit at the cost of increased computational complexity during the matching process. In our tests we use $M=30$. The additional term \( S(x_i, x_j) \) captures the direct similarity between the two images being evaluated. This is weighted by the hyperparameter $\alpha>0$ which can be tuned on the training set comparisons. In our case, we use the default hyperparameter setting $\alpha=1$. This ensures that the Relative Similarity Metric reflects not only how similarly the probe image \( x_i \) and candidate image \( x_j \) relate to all other samples in the dataset, but also includes their own mutual similarity directly in the final score. Note that the higher the RSM value, the more dissimilar the two fixed images are from one another.

This metric is particularly powerful in hashing-based systems where the binary nature of embeddings limits the resolution of direct similarity scores. RSM provides an additional layer of discriminative power by incorporating structural information across the dataset.

To our knowledge, no prior work in palmprint recognition leverages this type of relational similarity comparison. While similar relational insights have been explored in contrastive learning for fingerprint and face recognition \cite{tang2024intrafinger, hermans2017defense}, our application of RSM to palmprint matching presents a novel contribution.

\section{Results}

To evaluate the effectiveness of the proposed Relative Similarity Metric (RSM), we benchmarked its performance on the Tongji palmprint dataset. Table~1 summarizes the Equal Error Rates (EERs) achieved by a range of baseline and state-of-the-art palmprint recognition methods, including traditional handcrafted feature approaches, deep feature learning, and hashing-based systems.

The Tongji palmprint dataset \cite{zhang2017towards} contains 12,000 contactless palmprint images from 300 individuals (600 palms), acquired using a custom-designed non-contact capture device. Each palm was imaged in two separate sessions, with ten images per palm collected per session, resulting in 20 images per individual. This structure enables robust evaluation of both intra-session and inter-session variability, simulating real-world deployment conditions. In total, the dataset includes 60,000 genuine match comparisons and 35,940,000 impostor match comparisons, making it a strong benchmark for evaluating performance under severe class imbalance and fine-grained identity separation. All other methods—including our proposed variant—were trained using only the first session and evaluated on the second session.

\begin{table}
\centering
\caption{EERs (\%) on the Tongji Dataset}
\begin{tabular}{l c}
\toprule
\textbf{Method} & \textbf{EER (\%)} \\
\midrule
PalmCode & 0.1100 \\
Ordinal Code & 0.1600\\
Fusion Code & 0.0731 \\
CompCode & 0.1100 \\
RLOC & 0.0253 \\
HOC & 0.0954 \\
DOC & 0.0431 \\
DCC & 0.0506 \\
DRCC & 0.0308 \\
BOCV & 0.0056 \\
E-BOCV & 0.0180 \\
EDM & 0.0113 \\
2TCC & 0.0075 \\
MTCC & 0.0043 \\
DHN & 0.0879 \\
DHPN & 0.0659 \\
PalmNet & 0.0322 \\
CompNet & 0.0250 \\
CCNet & 0.000042 \\
\textbf{Ours (CCNet + RSM)} & \textbf{\underline{0.000036}} \\
\bottomrule
\end{tabular}
\end{table}

As shown in Table~1, earlier methods such as PalmCode, Ordinal Code, and Fusion Code produce EERs in the range of 0.07–0.16\%. Deep learning-based methods such as PalmNet (0.0322\%), DHPN (0.0659\%), and DHN (0.0879\%) significantly reduce the EER, leveraging learned feature representations.

Among recent state-of-the-art approaches, CCNet achieves an exceptionally low EER of 0.000042\% by utilizing a robust feature encoder and cosine similarity-based matching. However, cosine similarity only captures the relationship between the two images being compared, without considering the structure of similarity relationships across the entire dataset.

Our approach enhances CCNet by integrating the proposed Relative Similarity Metric (RSM), which measures the contextual consistency of similarity values with respect to the entire gallery. This subtle but powerful shift enables the model to better distinguish true matches from outliers and impostor pairs. As a result, our method achieves a new state-of-the-art EER of 0.000036\% on the Tongji dataset.

This improvement, albeit marginal in absolute terms, is significant in highly saturated operating regimes where even a handful of incorrect matches can be critical (here we have a relative improvement of 15\% compared to the CCNet model using the cosine distance metric for matching). It demonstrates the power of embedding relational structure into the matching process, allowing for better separation and confidence in final match decisions. On top of this, the hyperparameter $\alpha$ was set to $1$ for this experiment (the default setting). Careful tuning of this parameter is expected to yield better performance.

\begin{figure}
    \centering
    \includegraphics[width=0.49\textwidth]{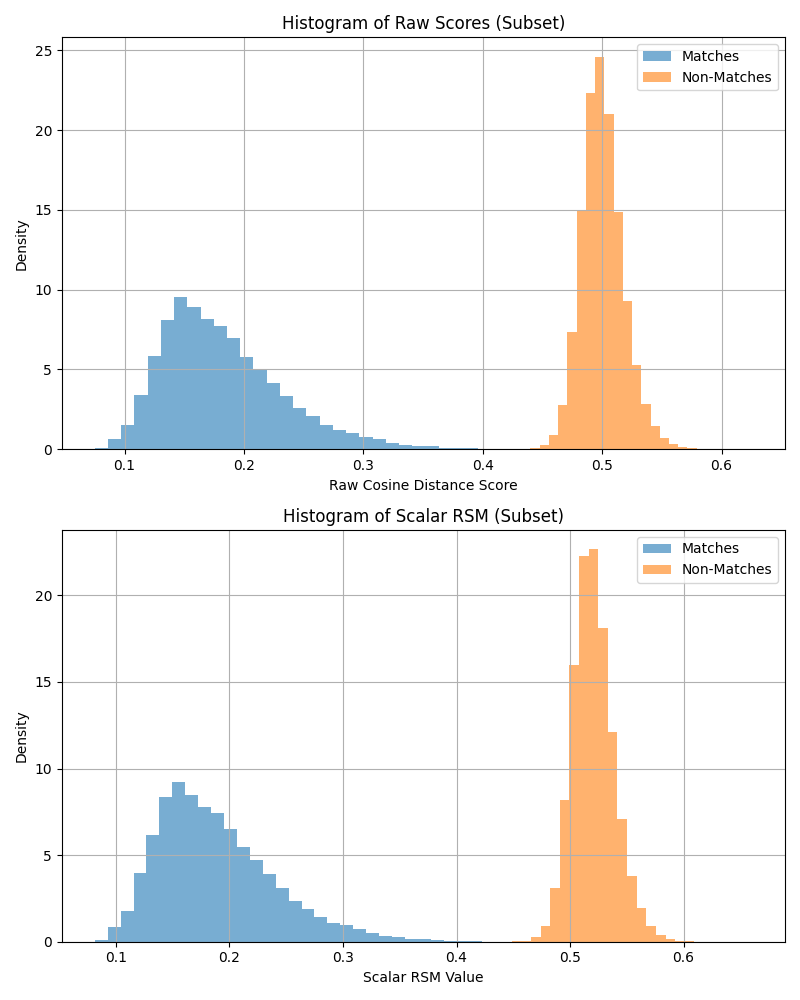}
    \caption{Comparison of score distributions for match and non-match pairs using cosine similarity versus the proposed Relative Similarity Metric (RSM).}
    \label{fig:histogram_comparison_trouble}
\end{figure}

\begin{figure}[h]
    \centering
    \includegraphics[width=0.49\textwidth]{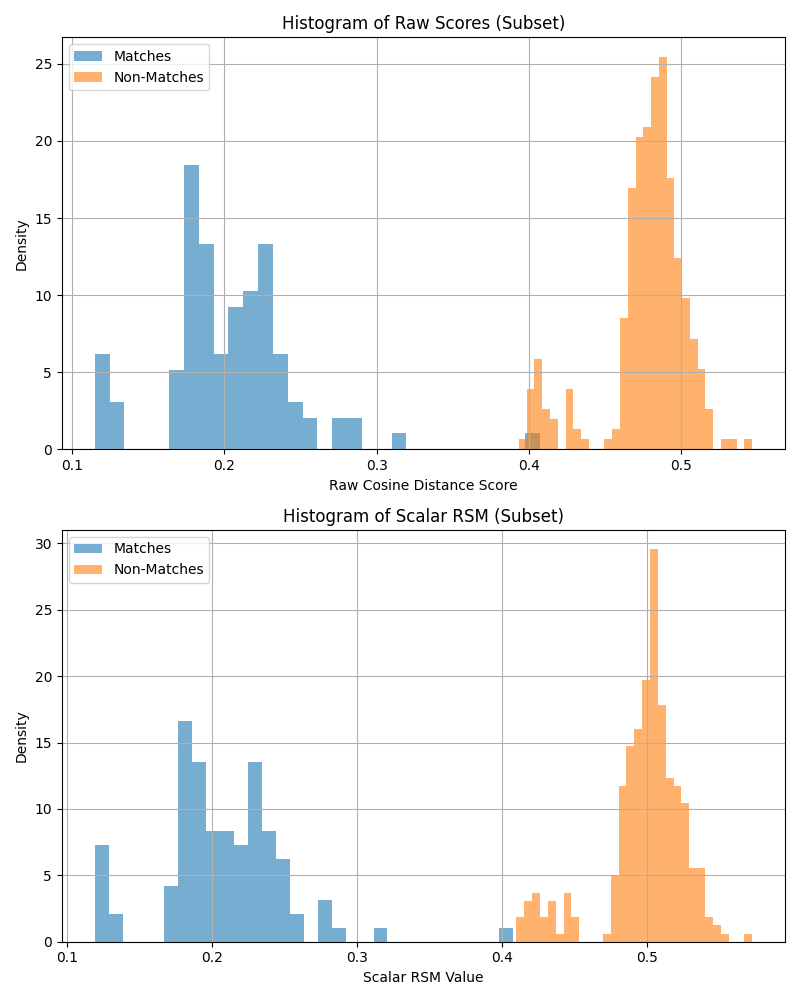}
    \caption{Comparison of a subset of 'difficult' score distributions for match and non-match pairs using cosine similarity versus the proposed Relative Similarity Metric (RSM).}
    \label{fig:histogram_comparison}
\end{figure}

Figure~\ref{fig:histogram_comparison} shows the distribution of match and non-match scores using both cosine similarity and the proposed Relative Similarity Metric. While the overall separation between the two classes remains consistent across both approaches, a key difference is the suppression of anomalous scores under RSM. In particular, the Relative Similarity Metric reduces the occurrence of unusually high similarity scores among impostor pairs and unusually low similarity scores among genuine pairs. This regularization effect results in fewer overlapping scores near the decision threshold, which directly accounts for the observed reduction in EER.

In addition to the histogram containing all match and non-match comparisons, we include a second figure (Figure \ref{fig:histogram_comparison_trouble}) that focuses exclusively on users associated with errors under the cosine distance metric — the so-called trouble comparisons. These are cases where the cosine distance score incorrectly suggests a match or non-match.

While the first figure provides an overview of the overall score distributions, it is dominated by 'easy' or correctly classified comparisons, making it difficult to observe the overlap between matches and non-matches in problematic regions. In contrast, the second figure isolates only the confusing cases, and reveals that the Relative Similarity Metric (RSM) yields better separation — with minimal to no overlap between the two histograms. This demonstrates RSM's improved ability to distinguish between genuine matches and impostor pairs in scenarios where cosine distance fails.

\section{Conclusion}

In this paper, we proposed the Relative Similarity Metric (RSM) for palmprint recognition that complements existing deep hashing and classification-based architectures by introducing contextual awareness into the matching process. Unlike traditional pairwise similarity metrics such as cosine distance, which consider only the direct relationship between two images, RSM evaluates how consistently one image relates to the rest of the dataset—providing a relational, second-order signal that enhances match reliability.

We applied RSM on top of the state-of-the-art CCNet architecture and demonstrated its effectiveness on the Tongji dataset. Our method achieved a new state-of-the-art Equal Error Rate (EER) of 0.000036\%, improving upon CCNet’s already competitive baseline of 0.000042\%. This performance gain stems from RSM’s ability to suppress spurious high impostor similarities and reduce false negatives in genuine matches, thereby tightening the decision boundary.

This work opens a new direction in palmprint recognition by emphasizing relational matching metrics over isolated similarity scores. Future work will explore tuning techniques for the hyperparameter in the RSM calculation, the integration of RSM into the learning objective itself, as well as its application to other biometric modalities and large-scale retrieval tasks.



\begin{thebibliography}{99}

\bibitem{svoboda2016palmprint}
J. Svoboda, M. Humenberger, and M. Zeppelzauer.
\newblock Palmprint recognition via discriminative index learning.
\newblock In \emph{2016 23rd International Conference on Pattern Recognition (ICPR)}, pages 1508--1513. IEEE, 2016.
\newblock 

\bibitem{liu2023smhnet}
X. Liu, B. Qin, and Q. Chen.
\newblock SMHNet: Similarity Metric Hashing Network for Few-Shot Palmprint Recognition.
\newblock \emph{IEEE Access}, vol. 11, pp. 20304--20316, 2023.
\newblock 

\bibitem{tang2024intrafinger}
Y. Tang, H. Zhang, and L. Tian.
\newblock Unveiling Intra-Person Fingerprint Similarity via Deep Contrastive Learning.
\newblock \emph{Patterns}, 2024.
\newblock 

\bibitem{ccnet2023}
Z. Yang, H. Huangfu, L. Leng, B. Zhang, A. B. J. Teoh, and Y. Zhang.
\newblock Comprehensive Competition Mechanism in Palmprint Recognition.
\newblock \emph{IEEE Transactions on Information Forensics and Security}, vol. 18, pp. 5160--5167, 2023.
\newblock 

\bibitem{deepdistillation2021}
H. Shao, D. Zhong, and X. Du.
\newblock Deep Distillation Hashing for Unconstrained Palmprint Recognition.
\newblock \emph{IEEE Transactions on Instrumentation and Measurement}, vol. 70, pp. 1--13, 2021.
\newblock 

\bibitem{zhang2020palmnet}
J. Zhang, Y. Ma, and M. K. Sun.
\newblock PalmNet: A Deep Network for Palmprint Identification and Verification.
\newblock \emph{Neurocomputing}, vol. 412, pp. 1--10, 2020.
\newblock 

\bibitem{kong2006competitive}
A. W.-K. Kong, D. Zhang, and M. Kamel.
\newblock A survey of palmprint recognition.
\newblock Pattern Recognition, vol. 39, no. 3, pp. 311–328, 2006.

\bibitem{zhang2003online}
D. Zhang, W. Kong, J. You, and M. Wong.
\newblock Online palmprint identification.
\newblock IEEE Transactions on Pattern Analysis and Machine Intelligence, vol. 25, no. 9, pp. 1041–1050, 2003.

\bibitem{wu2004palmprint}
X. Wu, K. Wang, D. Zhang, and B. Huang.
\newblock Palmprint classification using principal lines.
\newblock Pattern Recognition, vol. 37, no. 10, pp. 1987–1998, 2004.

\bibitem{hermans2017defense}
A. Hermans, L. Beyer, and B. Leibe.
\newblock In defense of the triplet loss for person re-identification.
\newblock arXiv preprint arXiv:1703.07737, 2017.

\bibitem{yang2018scalable}
X. Yang, S. Wang, M. Wang, and H. Tao.
\newblock Scalable Deep Hashing with Region Awareness.
\newblock IEEE Transactions on Image Processing, vol. 27, no. 10, pp. 4835–4848, 2018.

\bibitem{lai2015simultaneous}
H. Lai, Y. Pan, Y. Liu, and S. Yan.
\newblock Simultaneous feature learning and hash coding with deep neural networks.
\newblock In CVPR, pp. 3270–3278, 2015.

\bibitem{wang2017deep}
J. Wang, T. Zhang, J. Song, N. Sebe, and H. T. Shen.
\newblock A Survey on Learning to Hash.
\newblock IEEE Transactions on Pattern Analysis and Machine Intelligence, vol. 40, no. 4, pp. 769–790, 2017.

\bibitem{li2015feature}
W. Li, Y. Zhao, J. Chen, and S. Member.
\newblock Feature selection and weighted subspace learning for face recognition.
\newblock Pattern Recognition, vol. 48, no. 12, pp. 3846–3856, 2015.

\bibitem{zhang2021comprehensive}
B. Zhang, Y. Zhang, Z. Yang, and L. Leng.
\newblock A Comprehensive Palmprint Recognition Framework Using Adaptive Loss.
\newblock IEEE Access, vol. 9, pp. 117407–117417, 2021.

\bibitem{zhang2017towards}
L.~Zhang, L.~Li, A.~Yang, Y.~Shen, and M.~Yang,
``Towards contactless palmprint recognition: A novel device, a new benchmark, and a collaborative representation based identification approach,''
\emph{Pattern Recognition}, vol. 69, pp. 199--212, Sep. 2017.

\bibitem{zhong2019handgraph}
D. Zhong, H. Shao, and X. Du.
\newblock A hand-based multi-biometrics via deep hashing network and biometric graph matching.
\newblock \emph{IEEE Transactions on Information Forensics and Security}, vol. 14, no. 12, pp. 3140--3150, Dec. 2019.

\bibitem{zhong2019survey}
D. Zhong, X. Du, and K. Zhong.
\newblock Decade progress of palmprint recognition: A brief survey.
\newblock \emph{Neurocomputing}, vol. 328, pp. 16--28, Feb. 2019.

\bibitem{zhang2018lbp}
S. Zhang, H. Wang, W. Huang, and C. Zhang.
\newblock Combining modified LBP and weighted SRC for palmprint recognition.
\newblock \emph{Signal, Image and Video Processing}, vol. 12, no. 6, pp. 1035--1042, Sep. 2018.

\bibitem{fei2022jointly}
L. Fei, B. Zhang, Y. Xu, C. Tian, I. Rida, and D. Zhang.
\newblock Jointly heterogeneous palmprint discriminant feature learning.
\newblock \emph{IEEE Transactions on Neural Networks and Learning Systems}, vol. 33, no. 9, pp. 4979--4990, Sep. 2022.

\bibitem{fei20203dpalm}
L. Fei, B. Zhang, W. Jia, J. Wen, and D. Zhang.
\newblock Feature extraction for 3-D palmprint recognition: A survey.
\newblock \emph{IEEE Transactions on Instrumentation and Measurement}, vol. 69, no. 3, pp. 645--656, Mar. 2020.

\bibitem{genovese2019palmnet}
A. Genovese, V. Piuri, K. N. Plataniotis, and F. Scotti.
\newblock PalmNet: Gabor-PCA convolutional networks for touchless palmprint recognition.
\newblock \emph{IEEE Transactions on Information Forensics and Security}, vol. 14, no. 12, pp. 3160--3174, Dec. 2019.

\bibitem{chen2019lgcn}
P. Chen, W. Li, L. Sun, X. Ning, L. Yu, and L. Zhang.
\newblock LGCN: Learnable Gabor convolution network for human gender recognition in the wild.
\newblock \emph{IEICE Transactions on Information and Systems}, vol. 102, no. 10, pp. 2067--2071, 2019.

\bibitem{zhong2020cosine}
D. Zhong and J. Zhu.
\newblock Centralized large margin cosine loss for open-set deep palmprint recognition.
\newblock \emph{IEEE Transactions on Circuits and Systems for Video Technology}, vol. 30, no. 6, pp. 1559--1568, Jun. 2020.

\bibitem{wu2021fusion}
T. Wu, L. Leng, M. K. Khan, and F. A. Khan.
\newblock Palmprint-palmvein fusion recognition based on deep hashing network.
\newblock \emph{IEEE Access}, vol. 9, pp. 135816--135827, 2021.

\bibitem{dong2022cohash}
X. Dong, M. K. Khan, L. Leng, and A. B. J. Teoh.
\newblock Co-learning to hash palm biometrics for flexible IoT deployment.
\newblock \emph{IEEE Internet of Things Journal}, vol. 9, no. 23, pp. 23786--23794, Dec. 2022.

\bibitem{zhu2016dhn}
H. Zhu, M. Long, J. Wang, and Y. Cao,
\newblock Deep hashing network for efficient similarity retrieval,
\newblock in \emph{Proc. AAAI Conf. Artif. Intell. (AAAI)}, vol. 30, no. 1, 2016, pp. 1--7.

\bibitem{khosla2020supervised}
P. Khosla, P. Teterwak, C. Wang, A. Sarna, Y. Tian, P. Isola, A. Maschinot, C. Liu, and D. Krishnan,
\newblock Supervised contrastive learning,
\newblock in \emph{Advances in Neural Information Processing Systems (NeurIPS)}, vol. 33, pp. 18661--18673, 2020.

\bibitem{fei2016half}
L. Fei, Y. Xu, and D. Zhang,
\newblock Half-orientation extraction of palmprint features,
\newblock \emph{Pattern Recognition Letters}, vol. 69, pp. 35--41, Jan. 2016.

\bibitem{guo2009bocv}
Z. Guo, D. Zhang, L. Zhang, and W. Zuo,
\newblock Palmprint verification using binary orientation co-occurrence vector,
\newblock \emph{Pattern Recognition Letters}, vol. 30, no. 13, pp. 1219--1227, Oct. 2009.

\bibitem{yang2021extreme}
Z. Yang, L. Leng, and W. Min,
\newblock Extreme downsampling and joint feature for coding-based palmprint recognition,
\newblock \emph{IEEE Transactions on Instrumentation and Measurement}, vol. 70, pp. 1--12, 2021.

\bibitem{yang2023multiorder}
Z. Yang, L. Leng, T. Wu, M. Li, and J. Chu,
\newblock Multi-order texture features for palmprint recognition,
\newblock \emph{Artificial Intelligence Review}, vol. 56, no. 2, pp. 995--1011, Feb. 2023.

\bibitem{kong2006fusion}
A. Kong, D. Zhang, and M. Kamel,
\newblock Palmprint identification using feature-level fusion,
\newblock \emph{Pattern Recognition}, vol. 39, no. 3, pp. 478--487, Mar. 2006.


\end{thebibliography}
\end{document}